\documentclass[letterpaper]{article} 
\usepackage{aaai2026}  
\usepackage{times}  
\usepackage{helvet}  
\usepackage{courier}  
\usepackage[hyphens]{url}  
\usepackage{graphicx} 
\usepackage{natbib}  
\usepackage{caption} 
\usepackage{algorithm}
\usepackage{algorithmic}

\usepackage{newfloat}
\usepackage{listings}
\DeclareCaptionStyle{ruled}{labelfont=normalfont,labelsep=colon,strut=off} 
\floatstyle{ruled}
\newfloat{listing}{tb}{lst}{}
\floatname{listing}{Listing}
\title{Binary-Gaussian: Compact and Progressive Representation for 3D Gaussian Segmentation}
\author{
    An Yang\textsuperscript{\rm 1}, Chenyu Liu\textsuperscript{\rm 2}, Jun Du\textsuperscript{\rm 1}\thanks{Corresponding author.}, Jianqing Gao\textsuperscript{\rm 2}, Jia Pan\textsuperscript{\rm 2}, Jinshui Hu\textsuperscript{\rm 2}, Baocai Yin\textsuperscript{\rm 2}, Bing Yin\textsuperscript{\rm 2}, Cong Liu\textsuperscript{\rm 2}
}
\affiliations{
    \textsuperscript{\rm 1}NERC-SLIP, University of Science and Technology of China\\
    \textsuperscript{\rm 2}iFLYTEK Research\\

    yangan2002@mail.ustc.edu.cn, jundu@ustc.edu.cn, \{cyliu7,jqgao,jiapan,jshu,bcyin,bingyin,congliu2\}@iflytek.com\\
}

\usepackage{bibentry}
\usepackage{booktabs} 
\usepackage{multirow}
\usepackage{amsmath}
\usepackage{amssymb}
\usepackage{enumitem}
\usepackage{multirow}     
\usepackage{pifont}
\begin{document}

\maketitle

\begin{abstract}
3D Gaussian Splatting (3D-GS) has emerged as an efficient 3D representation and a promising foundation for semantic tasks like segmentation. However, existing 3D-GS-based segmentation methods typically rely on high-dimensional category features, which introduce substantial memory overhead. Moreover, fine-grained segmentation remains challenging due to label space congestion and the lack of stable multi-granularity control mechanisms.
To address these limitations, we propose a coarse-to-fine binary encoding scheme for per-Gaussian category representation, which compresses each feature into a single integer via the binary-to-decimal mapping, drastically reducing memory usage. We further design a progressive training strategy that decomposes panoptic segmentation into a series of independent sub-tasks, reducing inter-class conflicts and thereby enhancing fine-grained segmentation capability.
Additionally, we fine-tune opacity during segmentation training to address the incompatibility between photometric rendering and semantic segmentation, which often leads to foreground-background confusion.
Extensive experiments on multiple benchmarks demonstrate that our method achieves state-of-the-art segmentation performance while significantly reducing memory consumption and accelerating inference.
\end{abstract}

\section{Introduction}
Understanding and manipulating 3D scenes---including representation, rendering, perception, and editing---is a long-standing goal in computer vision and graphics, supporting applications such as virtual reality, robotics, and digital twins.
3D Gaussian Splatting(3D-GS)~\cite{kerbl20233d}, with its discrete spatial structure and fast rasterization, is becoming a mainstream approach for 3D scene representation, offering significant advantages in real-time interaction~\cite{choi2024click,zhao2025isegman} and scene editing~\cite{yan20243dsceneeditor,gaussian_grouping,qu2025drag}.
For such applications, accurately and efficiently segmenting objects is important~\cite{jiang2024vr}. However, this remains a challenging problem, especially in complex environments where objects exhibit occlusion, varying scales, and weak boundaries.

Recently, various segmentation methods~\cite{gaussian_grouping,shen2024flashsplat,choi2024click,cen2025segment} based on 3D-GS have been proposed. To represent category information, each Gaussian is usually assigned an additional continuous-valued vector. These approaches learn per-Gaussian category features from multi-view 2D mask annotations using techniques such as global optimization or contrastive learning.
During inference, segmentation is achieved either by taking the argmax over class probabilities, or by performing clustering in the feature space.
However, the additional feature vector assigned to each Gaussian (typically 32 dimensions) accounts for over 50\% of the original parameter size (62 dimensions), resulting in significant memory overhead. 
In addition, methods that support multi-granularity segmentation, such as SAGA~\cite{cen2025segment}, control the segmentation scale by simply multiplying a scale-controlling embedding with the category features. However, this approach is unstable and struggles to deliver accurate results for fine-grained segmentation.

To alleviate these issues, we first replace continuous category features with binary codes. By leveraging the one-to-one mapping between binary vectors and decimal integers, each Gaussian’s category feature can be compactly stored as a single integer, significantly reducing the additional memory overhead introduced by segmentation.
We further adopt a coarse-to-fine multi-granularity representation and progressive training strategy. Specifically, the binary code is divided into multiple partitions, each encoding feature information for a specific granularity level. The segmentation at a finer level depends not only on the current partition but also on all coarser-level partitions. In this hierarchical structure, coarse-level features determine broad category assignments, while finer-level features further subdivide those categories---akin to a multi-level indexing scheme. This design decomposes the complex  segmentation task into a sequence of simpler, localized sub-problems, thereby reducing training complexity and significantly enhancing the model's ability to distinguish fine-grained categories.
Moreover, existing models typically inherit opacity from photometric reconstruction and keep it fixed, which can cause low-opacity foreground objects to be misclassified as background. This issue is amplified under discrete representations like binary encoding, so we fine-tune opacity to better align rendering behavior with semantic segmentation objectives.
Finally, our model achieves superior segmentation accuracy while significantly reducing storage overhead and accelerating inference, outperforming existing state-of-the-art methods across multiple benchmarks.
In summary, our main contributions are as follows:
\begin{itemize}
    \item We introduce binary codes to represent category features,  reducing the parameter overhead of segmentation to just 1.6\% and achieving over 700 FPS in segmentation rendering.
    \item We propose a coarse-to-fine multi-granularity representation and progressive training strategy, which simplifies the segmentation task and improves overall accuracy, especially for fine-grained segmentation.
    \item We identify the incompatibility between opacity in photometric and semantic reconstruction, and address it by fine-tuning opacity during segmentation training, which improves foreground-background separation.
    \item Extensive qualitative and quantitative experiments across diverse scenes demonstrate that our method consistently outperforms existing state-of-the-art approaches in segmentation quality.
\end{itemize}

\section{Related Work}
\paragraph{3D Segmentation in Radiance Fields.}
The integration of semantic segmentation into radiance field representations has become an active area of research, driven by the increasing demand for 3D scene understanding and object-level interaction. 
Initial work~\cite{kobayashi2022decomposing,tschernezki2022neural,goel2023interactive,kerr2023lerf} on 3D segmentation in radiance fields often relied on lifting visual features from 2D foundation models into 3D. Methods such as DFFs~\cite{kobayashi2022decomposing}, N3F~\cite{tschernezki2022neural}, and ISRF~\cite{goel2023interactive} adopt feature distillation pipelines, transferring representations from self-supervised 2D models like DINO~\cite{caron2021emerging} into 3D fields. However, since these models are not designed for segmentation, their transferred features tend to lack spatial precision. 
NeRF-SOS ~\cite{fan2022nerf} and ContrastiveLift~\cite{bhalgat2023contrastive}
distill 2D pixel or region level feature similarities into 3D representations, introducing contrastive learning as a new training paradigm.
Furthermore, some work~\cite{zhou2024feature,chen2023interactive} leverages the Segment Anything Model~\cite{kirillov2023segment}, distilling its encoder features into 3D Gaussians while using the decoder for segmentation inference. Despite improvements in segmentation quality, such methods often incur high computational costs.

Another line of work formulates 3D segmentation as a mask-lifting problem, where 2D masks are projected into 3D space across multiple views. Examples include SA3D~\cite{cen2023segment}, NVOS~\cite{ren2022neural}, MVSeg~\cite{mirzaei2023spin}, FlashSplat~\cite{shen2024flashsplat}, and GaussianGrouping~\cite{gaussian_grouping}, which rely on user prompts or video tracking to align masks and guide segmentation. While effective in certain settings, these approaches are vulnerable to view inconsistency and often assume ideal camera trajectories or tracking conditions.
More recent methods like OmniSeg3D~\cite{ying2023omniseg3d} and GARField~\cite{kim2024garfield} mitigate these issues by introducing contrastive learning schemes and scale-conditioned features, but they typically build on NeRF-based architectures, which are computationally expensive and not well-suited to real-time applications. 
Later works~\cite{choi2024click,cen2025segment} introduce contrastive learning into 3D Gaussians segmentation. Click-Gaussian~\cite{choi2024click} improves segmentation precision through coarse-to-fine global-guided learning, yet it is limited to dual-scale segmentation. SAGA~\cite{cen2025segment} introduces scale-gated affinity, similar to GARField, for adaptive granularity control, but its simple gating design-based on a single fully connected layer-limits precision and stability.
In contrast, our method employs a coarse-to-fine representation strategy to achieve stable multi-granularity segmentation.

\begin{figure*}[t]
\centering
\includegraphics[width=0.9\textwidth]{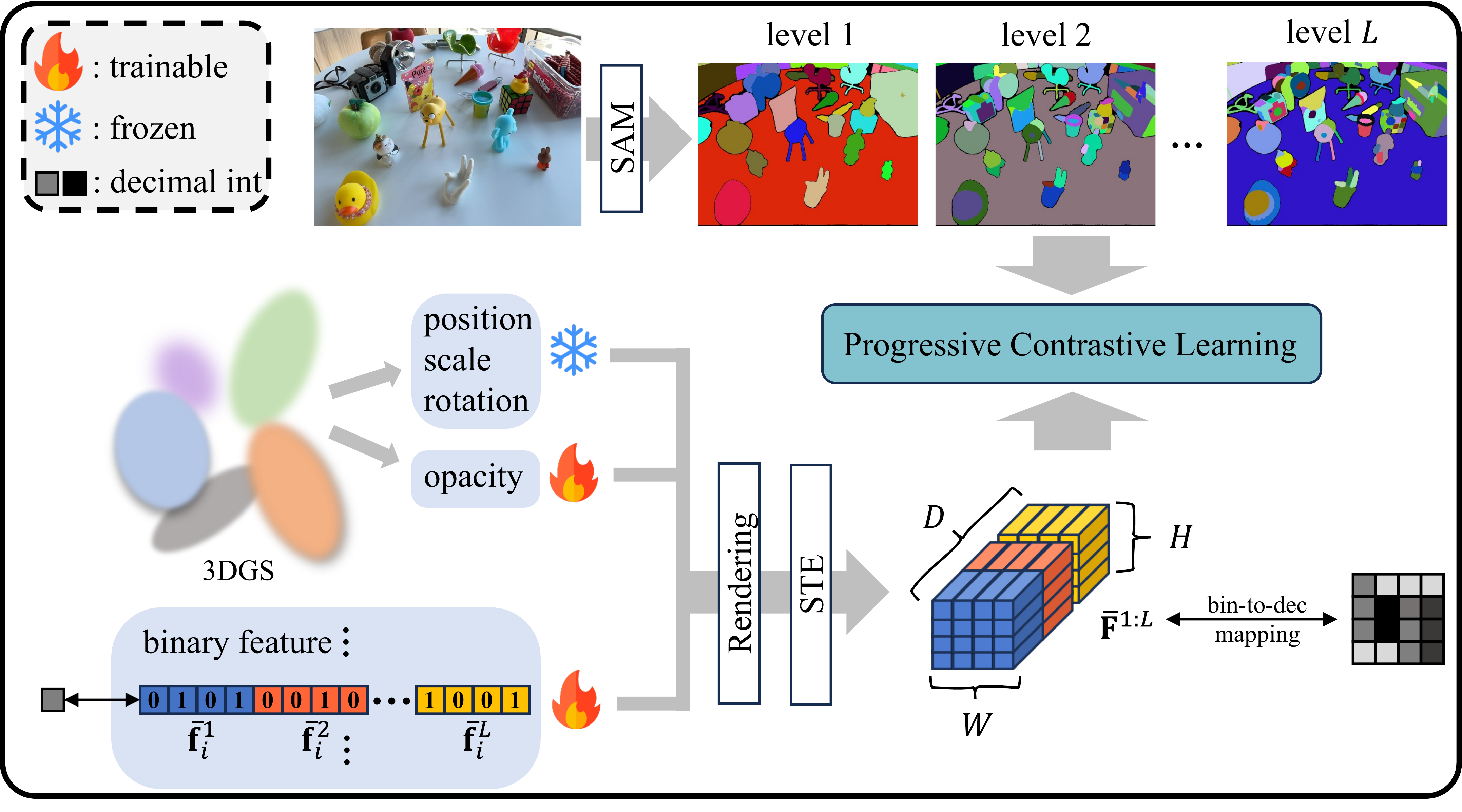} 
\caption{Overview of Binary-Gaussian. Our method builds upon pre-trained 3D Gaussians by augmenting each Gaussian with multi-granularity binary features. During training, we keep the opacity values learnable and optimize the multi-level binary features $\bar{\mathbf{F}}^{1:L}$ on the rendered 2D projections using a progressive contrastive learning strategy.}
\label{model1}
\end{figure*}

\paragraph{Compression for 3D Gaussian Splatting.}
3D-GS~\cite{kerbl3Dgaussians} enables real-time rendering with high quality, but suffers from high memory usage, as each Gaussian stores multiple attributes, often resulting in gigabyte-scale scenes and reduced rendering efficiency~\cite{fan2024lightgaussian}.  
To address this, recent methods compress 3D Gaussian Splatting by discretizing Gaussian attributes via codebook-based vector quantization.
CompGS~\cite{navaneet2023compact3d} reduces redundancy via opacity regularization and K-means quantization.  
LightGaussian~\cite{fan2024lightgaussian} integrates global significance-based pruning, SH distillation, and vector quantization into a unified compression framework.  
HAC~\cite{chen2024hac} introduces a hash-grid-assisted context model for entropy-aware compression and spatially consistent pruning.  
~\citet{papantonakis2024reducing} additionally prune resolution-insensitive Gaussians, adaptively reduce SH coefficients, and compress attributes via codebook quantization and half-precision representation.
Other efforts adopt either sensitivity-aware clustering with quantization-aware training~\cite{niedermayr2024compressed}, 
or grid-based color fields as a replacement for spherical harmonics~\cite{lee2024compact}, to reduce memory usage.
The above methods focus on compressing appearance-related parameters of 3D Gaussians. However, segmentation tasks require assigning additional attributes to each Gaussian to represent category information. Previous segmentation approaches typically rely on high-dimensional continuous features to encode such information, which significantly increases memory consumption. Unlike prior methods, our proposed binary feature encodes category semantics as compact binary codes, which can be stored using
an integer (e.g., a 32-bit integer), substantially reducing the storage overhead for 3D-GS-based segmentation models.

\section{Method}
\subsection{Preliminary: 3D Gaussian Splatting}
3D Gaussian Splatting(3D-GS) represents a 3D scene with a collection of explicit 3D Gaussians and renders images through a differentiable rasterizer~\cite{kerbl3Dgaussians}.
Given a training set of images \( \mathcal{I} = \{ I^v \}_{v=1}^V \) with camera poses, 3D-GS learns a set of 3D Gaussians \( G = \{ g_i \}_{i=1}^N \),  
where \( V \) is the number of views and \( N \) denotes the total number of Gaussians.  
Each Gaussian \( g_i \) is parameterized by its 3D position and covariance, opacity, and color represented using spherical harmonics coefficients.
Given a specific camera pose, the color of pixel $p$ is computed by projecting the Gaussians onto the 2D image plane and blending a depth-ordered subset \( \mathcal{N} \) overlapping the pixel $p$:
\begin{equation}
    C_p = \sum_{i \in \mathcal{N}} c_i \alpha_i T_i,
\end{equation}
where \( c_i \) is the color of each Gaussian and \( \alpha_i \) is given by evaluating a 2D Gaussian with covariance $\Sigma$ multiplied with a
learned per-Gaussian opacity~\cite{kerbl3Dgaussians},
and \( T_i = \prod_{j=1}^{i-1} (1 - \alpha_j) \) is the transmittance.

\subsection{Coarse-to-Fine Binary Feature}
Different from previous methods that adopt continuous values, we equip each 3D Gaussian with a binary code to represent its segmentation category. To support multi-granularity segmentation, we further introduce a coarse-to-fine representation scheme, where each Gaussian is associated with a set of hierarchical codes that capture segmentation labels at different levels of granularity. 

Specifically, let $L$ denote the maximum level of segmentation granularity. We assign $L$ feature vectors $\{ \bar{\mathbf{f}}_i^l \in \{0, 1\}^{D_l} \mid l = 1,2,\dots,L\}$ to each Gaussian to represent features at different levels, which are then concatenated into a single feature vector $\bar{\mathbf{f}}_{i} \in \mathbb{R}^{D}$ for efficient rendering computation, where $D = \sum^L_{l=1} D_l$. Using the rasterizer, we can render the features of different granularity levels to the 2D pixel $p$:
\begin{equation}
    \mathbf{F}_p = \sum_{i \in \mathcal{N}} \bar{\mathbf{f}_{i}} \alpha_{i} T_{i},
    \label{rendering}
\end{equation}
Then, we apply a straight-through estimator (STE) to obtain binary features:
\begin{equation}
    \bar{\mathbf{F}}_p = \text{stop\_grad} \left( \mathbb{I}(\mathbf{F}_p > 0.5) - \mathbf{F}_p \right) + \mathbf{F}_p
\end{equation}
where $\mathbb{I}(\cdot)$ is the element-wise indicator function that outputs 1 if the input condition is true and 0 otherwise, and $\text{stop\_grad}(\cdot)$ represents the stop-gradient operation that blocks gradients during backpropagation. $\bar{\mathbf{F}}_p \in \{0,1\}^{D}$ is the final binarized feature vector, and $\bar{\mathbf{F}}_p^l = \bar{\mathbf{F}}_p [ \sum^{l-1}_{k=1} D_k : \sum^{l}_{k=1} D_k ]$ denotes the binary features corresponding to the $l$-th level, with $l=1,2,\dots,L$.

The coarse-to-fine design is realized by progressively aggregating features across granularity levels. 
Specifically, for each granularity level $l$, we concatenate the binary feature vectors from all levels less than or equal to $l$ to form a unified feature for segmentation:
\begin{equation}
\bar{\mathbf{F}}_p^{1:l} = \text{Concat}(\bar{\mathbf{F}}_p^1, \bar{\mathbf{F}}_p^2, \dots, \bar{\mathbf{F}}_p^l).
\end{equation}

\begin{figure}[t]
\centering
\includegraphics[width=\linewidth]{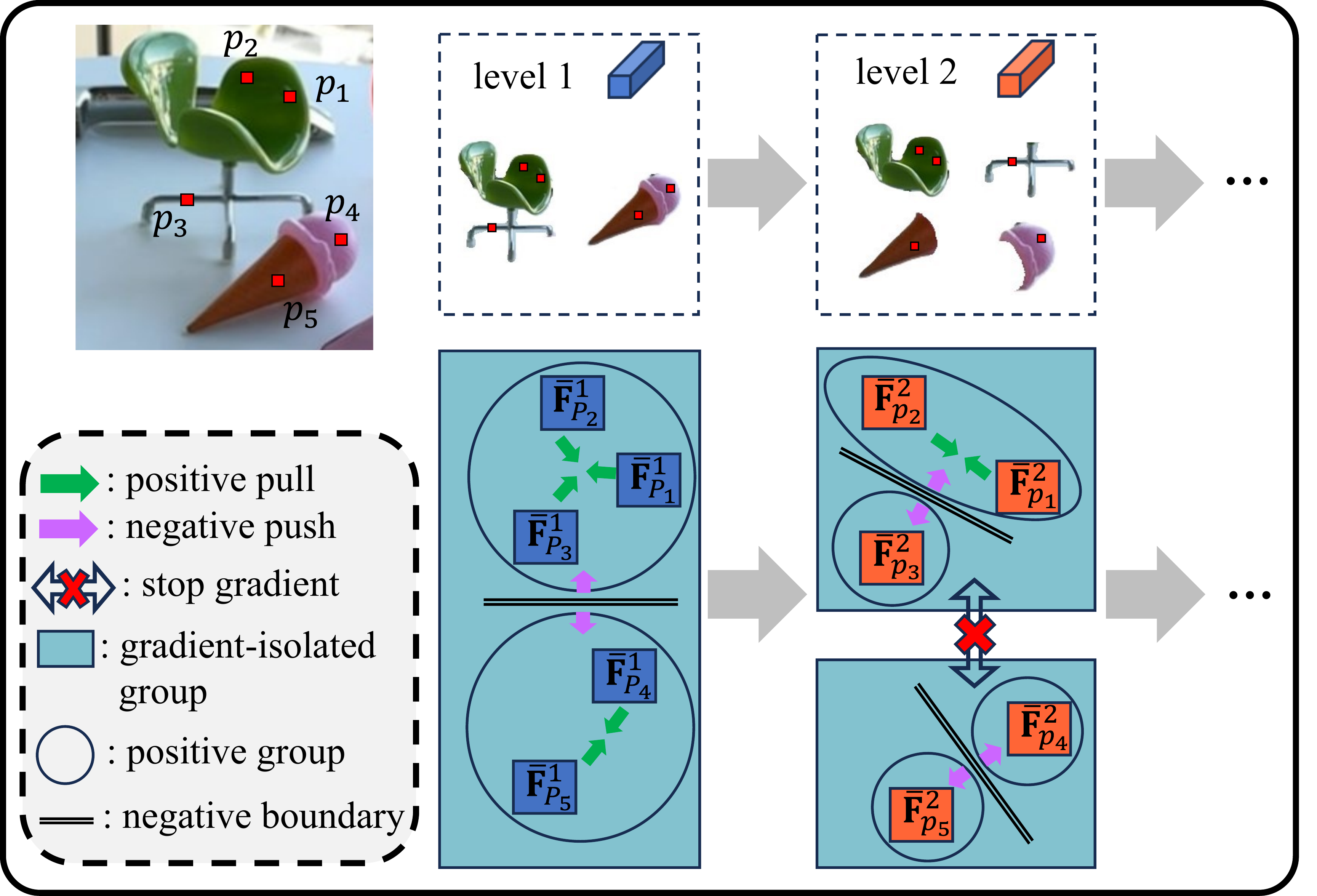} 
\caption{Illustration of Progressive Contrastive Learning. Fine-grained segmentation at each level is built on the coarser segmentation from the preceding level. Each positive group identified at the coarse level forms an independent, gradient-isolated sub-optimization task at the finer level.}
\label{model2}
\end{figure}

\subsection{Progressive Contrastive Learning}
We propose a novel progressive contrastive learning strategy tailored to multi-granularity binary features in a coarse-to-fine hierarchy. We next describe this strategy progressively. Let $M^l_p$ denote the mask that pixel $p$ belongs to at granularity level $l$.
\paragraph{Standard Contrastive Learning.}
We first adopt a standard contrastive learning approach for the level-$1$ features. For pixels $p$ and $q$, if they belong to the same level-$1$ mask, we minimize the L2 distance between their level-$1$ features. Otherwise, we maximize the L2 distance:
\begin{equation}
    \mathcal{L}_{\text{level-}1} =
    \begin{cases}
    \| \bar{\mathbf{F}}_p^1 - \bar{\mathbf{F}}_q^1 \|_2, &\text{if } M^1_p = M^1_q \\
    D_1 - \| \bar{\mathbf{F}}_p^1 - \bar{\mathbf{F}}_q^1 \|_2, &\text{otherwise.}
    \end{cases}
    \label{l1}
\end{equation}
Our objective is to ensure that $\bar{\mathbf{F}}_p^1 = \bar{\mathbf{F}}_q^1$ when $M^1_p = M^1_q$, and that $\bar{\mathbf{F}}_p^1 \ne \bar{\mathbf{F}}_q^1$ when $M^1_p \ne M^1_q$.

\paragraph{Progressive Contrastive Learning.}
Following the logic of mathematical induction, we assume that the level $l-1$ segmentation (with $l \ge 2$) has already been constrained. More specifically, the features $\bar{\mathbf{F}}^{1:l-1}$ are subject to constraints similar to Equation~\ref{l1}, ensuring that $\bar{\mathbf{F}}_p^{1:l-1} = \bar{\mathbf{F}}_q^{1:l-1}$ when $M^{l-1}_p = M^{l-1}_q$, and that $\bar{\mathbf{F}}_p^{1:l-1} \ne \bar{\mathbf{F}}_q^{1:l-1}$ when $M^{l-1}_p \ne M^{l-1}_q$.

For the segmentation at level $l$, we need to impose constraints to ensure that $\bar{\mathbf{F}}_p^{1:l} = \bar{\mathbf{F}}_q^{1:l}$ when $M^{l}_p = M^{l}_q$, and that $\bar{\mathbf{F}}_p^{1:l} \ne \bar{\mathbf{F}}_q^{1:l}$ when $M^{l}_p \ne M^{l}_q$.
if $M^{l}_p = M^{l}_q$, then it necessarily follows that $M^{l-1}_p = M^{l-1}_q$. According to our inductive assumption, constraints have already been applied such that $\bar{\mathbf{F}}_p^{1:l-1} = \bar{\mathbf{F}}_q^{1:l-1}$. Therefore, we only need to minimize the L2 distance between their level $l$ features $\bar{\mathbf{F}}_p^{l}$ and $\bar{\mathbf{F}}_q^{l}$, as shown by $p_1$ and $p_2$ at level 2 in Figure~\ref{model2}.
if $M^{l}_p \ne M^{l}_q$, we further consider two cases: 
\begin{equation*}
    (1) M^{l-1}_p \ne M^{l-1}_q; \qquad (2) M^{l-1}_p = M^{l-1}_q.
\end{equation*}
For case~(1), according to our inductive assumption, constraints have already been applied such that $\bar{\mathbf{F}}_p^{1:l-1} \ne \bar{\mathbf{F}}_q^{1:l-1}$.
Due to the inheritance property of the coarse-to-fine feature representation, it must hold that $\bar{\mathbf{F}}_p^{1:l} \ne \bar{\mathbf{F}}_q^{1:l}$.
Therefore, we do not impose inconsistent constraints on level $l$ features under this case, as shown by $p_1$ and $p_4$ at level 2 in Figure~\ref{model2}.
For case~(2), we have the constraint $\bar{\mathbf{F}}_p^{1:l-1} = \bar{\mathbf{F}}_q^{1:l-1}$. Hence, we maximize the L2 distance between their level $l$  features $\bar{\mathbf{F}}_p^{l}$ and $\bar{\mathbf{F}}_q^{l}$ to induce separation, resulting in $\bar{\mathbf{F}}_p^{1:l} \ne \bar{\mathbf{F}}_q^{1:l}$, as shown by $p_1$ and $p_3$ at level 2 in Figure~\ref{model2}.
Formally, we have:
\begin{equation}
\mathcal{L}_{\text{level-}l} =
\begin{cases}
\| \bar{\mathbf{F}}_p^l - \bar{\mathbf{F}}_q^l \|_2, 
& \text{if } M^l_p = M^l_q \\
 D_l - \| \bar{\mathbf{F}}_p^l - \bar{\mathbf{F}}_q^l \|_2 , 
& \text{else if } M^{l-1}_p = M^{l-1}_q\\
0, & \text{otherwise}
\end{cases}
\end{equation}
This completes our inductive reasoning. It demonstrates that by applying the set of constraints $\sum^L_{l=1} \mathcal{L}_{\text{level-}l}$ derived at each granularity level, we can optimize the multi-scale segmentation task. Notably, the above derivations for different cases rely on ground-truth labels, and there are no dependencies between features at different levels during training. As a result, all levels can be optimized simultaneously.
Our progressive contrastive learning decomposes the fine-grained segmentation task into a sequence of simpler sub-tasks, thereby reducing the complexity of the overall problem and significantly alleviating the overlap and confusion between features of different object categories.

\subsection{Virtual Negative Guidance for Positive-Only Contrast}
Contrastive learning depends on the equilibrium between positive attraction and negative repulsion. Disruptions to this balance have emerged as a key issue, motivating a wide range of research efforts to explore effective solutions~\cite{grill2020bootstrap,chen2021exploring,cha2023regularizing}.
In our progressive contrastive framework, this issue arises when a semantic mask becomes indivisible. At the next finer level, the resulting gradient-isolated group contains only positive pairs and receives no repulsive supervision. 
Counterintuitively, we observe that pull-only objectives in such cases often lead to feature drift or inconsistency, rather than feature collapse (as shown in Figure~\ref{virtual_negative} of ablation study).
To address this, we propose a simple yet effective solution: guiding feature learning using an all-one vector as the virtual negative.
Specifically, we impose an L2 regularization constraint on each pixel feature $\bar{\mathbf{F}}_p^{l}$ within a positive-only group at level $l$, aiming to maximize its distance from the virtual negative. Formally, we have:
\begin{equation}
\mathcal{L}_{\text{VN-}l} = \| \bar{\mathbf{F}}_p^l \|_2.
\end{equation}

\paragraph{Loss Function.}
In addition to the aforementioned contrastive loss, we introduce a regularization term on the pre-binarized features $\mathbf{F}_p$, encouraging them to approach binary values (0 or 1) for more semantically deterministic representations.
The regularization term is defined as:
\begin{equation}
\mathcal{L}_{\text{reg}} = \|\mathbb{I}(\mathbf{F}_p > 0.5) - \mathbf{F}_p\|_2^2
\end{equation}
The final loss function is defined as:
\begin{equation}
    \mathcal{L}_{\text{total}} = \lambda_{\text{reg}} \mathcal{L}_{\text{reg}} + \lambda_{\text{guiding}} \sum^L_l \mathcal{L}_{\text{VN-}l} + \lambda_{\text{con}} \sum^L_l \mathcal{L}_{\text{level-}l}
\end{equation}
To balance the magnitudes of different loss terms, we set $\lambda_{\text{reg}}=10$, $\lambda_{\text{guiding}}=1$, and $\lambda_{\text{con}}=1$.

\subsection{Additional Training Strategy}
\begin{figure}[t]
\centering
\includegraphics[width=\linewidth]{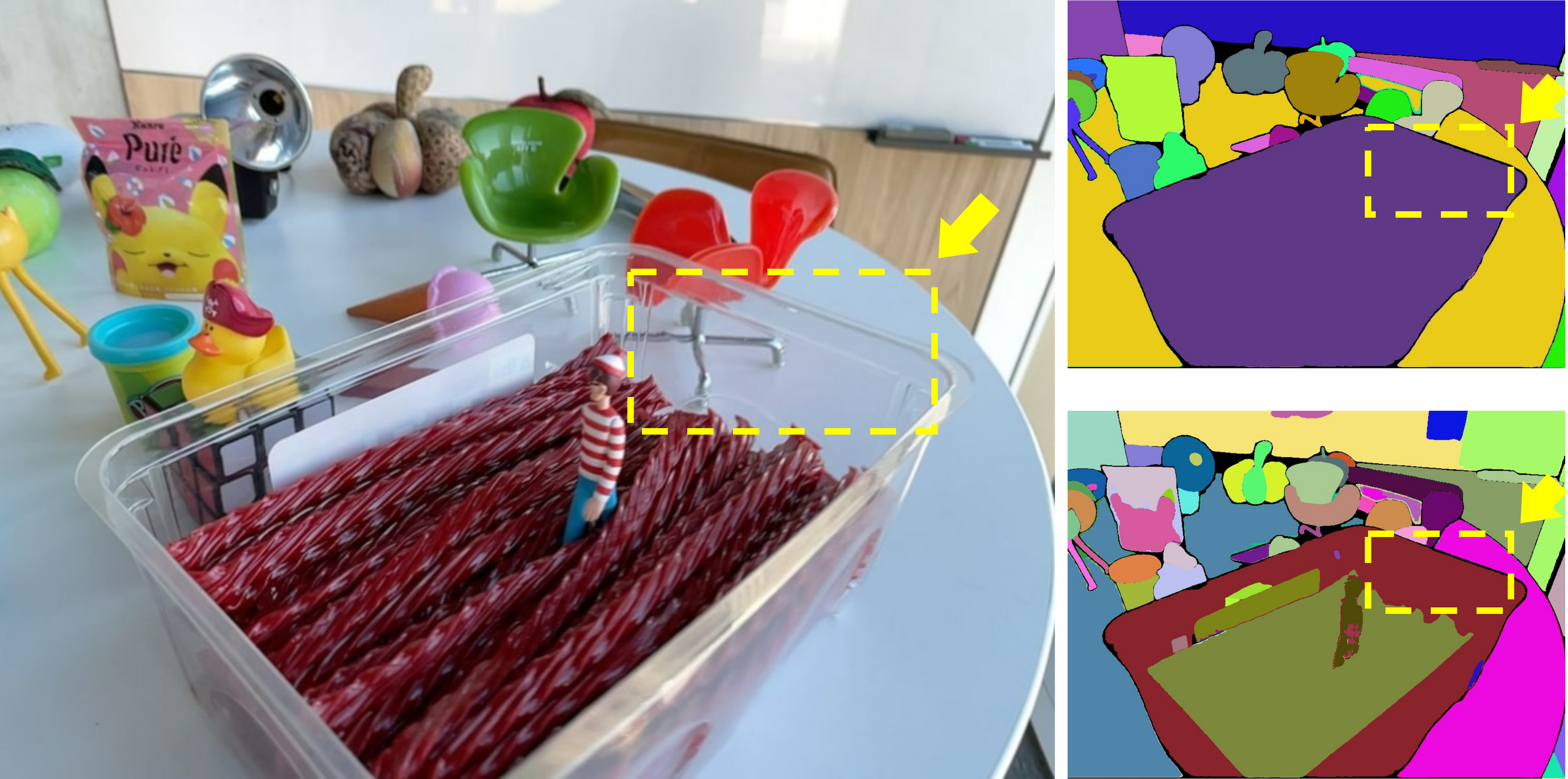} 
\caption{Opacity inconsistency between visual and segmentation rendering.}
\label{opacity}
\end{figure}

\paragraph{Counter-Visual Opacity Effects in Segmentation.}
Opacity, a fundamental attribute of each Gaussian, indicates its light-blocking strength and determines its contribution to the final pixel appearance through alpha blending.
In prior 3D Gaussian segmentation methods, opacity values are inherited from color reconstruction and kept fixed during training.
However, the opacity in color space does not always align with the semantics required for accurate segmentation.
A typical failure case arises when handling semi-transparent objects, such as a plastic box, as shown in Figure~\ref{opacity}. In the color space, the box is rendered with low opacity, allowing the background (e.g., a table) to be partially visible. 
However, when used in segmentation, low-opacity foreground Gaussians may contribute too little to dominate the pixel.
This issue is further exacerbated under discrete representations such as binary encoding. Unlike continuous features---which can compensate for low-opacity Gaussians by amplifying their feature values---our discrete encoding is bounded in range. As a result, low-opacity Gaussians are easily suppressed during feature rendering, causing insufficient optimization and foreground-background confusion. 
Therefore, we propose to fine-tune the opacity values during category feature training.

\paragraph{Mask-Balanced Sampling.}
In previous methods, each training iteration randomly samples $N_p$ pixels from an image to form $N_p \times N_p$ pixel pairs. This inevitably leads to data imbalance, where large masks occupying more pixels dominate the optimization, degrading segmentation performance for small masks~\cite{cen2025segment}.
To address this issue, we randomly sample only a portion of pixels from the entire image. For the remaining pixels, we adopt a Mask-Balanced Pixel Sampling strategy: we randomly select a set of masks, and then sample a fixed number of pixels from each selected mask.
This strategy increases the proportion of sampling pixels from small masks.

\subsection{Inference}
With the well-trained Gaussian binary features $\bar{\mathbf{f}}_i^{1:l}$, we can render the corresponding pixel-level binary feature map $\bar{\mathbf{F}}_p^{1:l}$ from arbitrary viewpoints. By applying binary-to-decimal mapping, we quickly obtain decimal integer labels for both $\bar{\mathbf{f}}_i^{1:l}$ and $\bar{\mathbf{F}}_p^{1:l}$:
\begin{equation*}
\text{Class}^l_{i} = \text{Bin2Dec}(\bar{\mathbf{f}}_i^{1:l}), \text{Class}^l_{p} = \text{Bin2Dec}(\bar{\mathbf{F}}_p^{1:l}),
\end{equation*}
where $\text{Bin2Dec}(\cdot)$ denotes the mapping function from binary to decimal. $\text{Class}^l_{i}$ and $\text{Class}^l_{p}$ indicate the semantic category at granularity level $l$, used for 3D Gaussian segmentation and multi-view 2D segmentation, respectively.
By associating these decimal values with each semantic object annotation, we can retrieve all Gaussians sharing the same specific value, thereby enabling object-specific 3D segmentation.

\section{Experiments}
\subsection{Datasets}
Following prior approaches~\cite{choi2024click,cen2025segment}, we employ a SAM-based automatic mask generation module to obtain segmentation annotations at multiple granularity levels within 3D scenes.
To evaluate our model's 3D segmentation performance, we adopt two widely used public real-world datasets: the LERF-Mask dataset~\cite{choi2024click} and the SPIn-NeRF dataset~\cite{mirzaei2023spin}. 
The LERF-Mask dataset comprises three scenes~\cite{kerr2023lerf} with manually annotated ground truth masks for both coarse and fine-grained segmentation.
The SPIn-NeRF dataset provides multi-view masks for single objects in the widely used NeRF datasets~\cite{fridovich2022plenoxels,knapitsch2017tanks,mildenhall2019local,mildenhall2021nerf,yen2022nerf}.
\subsection{Implementation Details}
We train the category features based on the 3D Gaussian parameters obtained from color-space reconstruction~\cite{kerbl20233d}. During this process, the learning rate for opacity is set to 0.001 (2\% of the initial value) for fine-tuning, while the learning rate for category features is set to 0.005. In each iteration, 2k pixels are randomly sampled, and an additional 8k pixels are selected using mask-balanced sampling. More training details are provided in Appendix A.

\begin{table*}[t]
\centering
\small
\begin{tabular}{l|cc|c|c|cc}
\toprule
\multirow{2}{*}{\textbf{Model}} 
& \multicolumn{2}{c|}{\textbf{Extra Param / Gau}} 
& \textbf{Multi-granularity}
& \multirow{2}{*}{\textbf{FPS}} 
& \multicolumn{2}{c}{\textbf{mIoU (\%)}} \\
& Size(bit) & Growth(\%) & \textbf{Support} &
& Coarse & Fine \\
\midrule
OmniSeg3D~\cite{ying2023omniseg3d}     & - & - & \ding{51}\ding{51}\ding{51} & 22.6 & 79.4 & 46.9 \\
GARField~\cite{kim2024garfield}     & - & - & \ding{51}\ding{51}\ding{51} & 0.32 & 80.9 & 71.4 \\
\midrule
Gau-Group~\cite{gaussian_grouping}     & 1024 & 51.6 & \ding{55} & \underline{288} & 72.8 & 31.5 \\
Feature3DGS~\cite{zhou2024feature}     & 4096 & 206.5 & \ding{55} & 14.6  & 65.6 & 63.5 \\
Click-Gaussian~\cite{choi2024click}     & 1024 & 51.6 & \ding{51} & $\approx 50\text{--}100$  & \textbf{89.1} & \underline{84.3} \\
SAGA~\cite{cen2025segment}             & 1024 & 51.6 & \ding{51}\ding{51}\ding{51} & 67.9  & 83.7 & 48.1 \\
SAGA$^{*}$~\cite{cen2025segment}       & \underline{32} & \underline{1.6} & \ding{51}\ding{51}\ding{51} & 74.6  & 69.7 & 47.8 \\
Ours                                   & \textbf{32} & \textbf{1.6} & \ding{51}\ding{51}\ding{51} & \textbf{769} & \underline{87.1} & \textbf{86.2} \\
\bottomrule
\end{tabular}
\caption{Comparison with baselines on LERF-Mask dataset in terms of parameter overhead, segmentation granularity, speed, and quality. 
The top two rows correspond to NeRF-based methods, and the bottom six rows to 3D-GS-based methods. SAGA* denotes the variant with parameter compression via codebook quantization. Symbol definitions: \ding{55} supports only a single segmentation granularity; \ding{51} supports two levels (coarse and fine); \ding{51}\ding{51}\ding{51} supports multi-granularity segmentation.}
\label{tab:lerf_summary}
\end{table*}

\subsection{Quantitative Evaluation}
We evaluate our method on the LERF-Mask dataset under the setting of multi-object, multi-granularity segmentation, with the results from a multi-perspective evaluation summarized in Table~\ref{tab:lerf_summary}. Firstly, leveraging the binary coding design, the category feature of each Gaussian occupies only 32 bits---merely 1.6\% of the original parameter size---significantly lower than competing methods. Moreover, while previous approaches rely on clustering to construct category centers and compute distances between Gaussian features and all centers during inference, our model simply maps the binary code to a decimal index for direct category identification. This enables inference speeds over 10× faster than prior methods, dramatically accelerating the rendering process.
Despite its lightweight and efficient design, our model also maintains high segmentation quality, achieving the best performance in fine-grained segmentation. The results on the SPIn‑NeRF dataset across diverse real-world scenes, as shown in Table~\ref{tab:spin_nerf}, further demonstrate the superiority of our method, which achieves state-of-the-art segmentation accuracy.

\begin{table}[t]
  \centering
  \begin{tabular}{l|c}
    \toprule
    Method & \textbf{mIoU (\%)} \\
    \midrule
    MVSeg~\cite{mirzaei2023spin}       & 90.9 \\
    SA3D~\cite{cen2023segment} & 92.4 \\
    SAGA~\cite{cen2025segment} & 88.0 \\
    Click‑Gaussian~\cite{choi2024click} & \underline{94.0} \\
    \textbf{Ours}             & \textbf{94.3} \\
    \bottomrule
  \end{tabular}
  \caption{Quantitative comparison with baselines on the SPIn‑NeRF dataset. We report the average mIoU across ten real-world scenes. Both MVSeg and SA3D are limited to segmenting a single object per training session.}
  \label{tab:spin_nerf}
\end{table}

\subsection{Qualitative Evaluation}
\begin{figure*}[!ht]
\centering
\includegraphics[width=\textwidth]{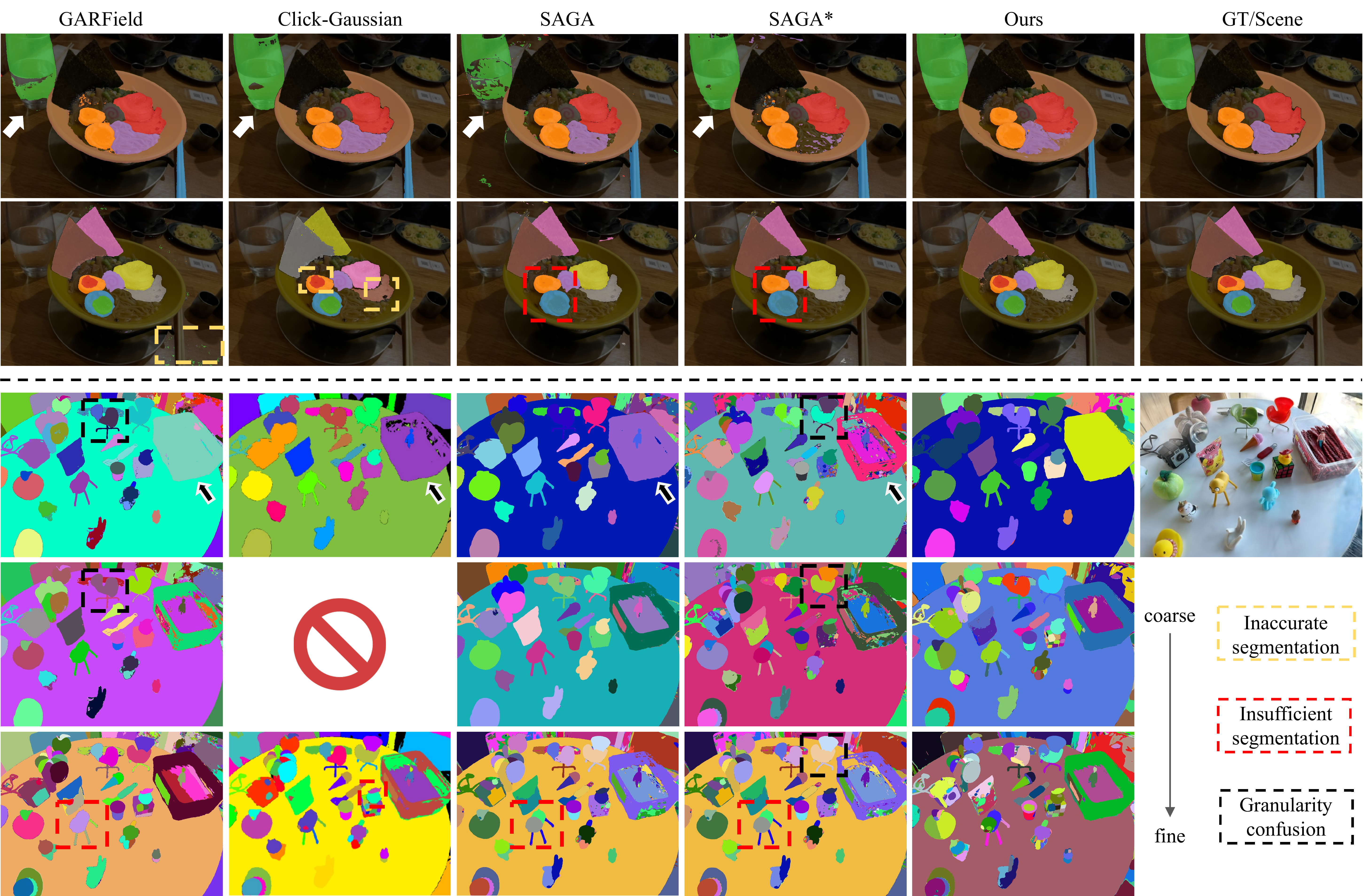} 
\caption{Comparison with baseline models on the LERF-Mask dataset. The top two rows illustrate object-specific segmentation results for coarse and fine-grained targets. The bottom three rows present panoptic segmentation results across different levels of granularity. The prohibition symbol indicates that Click-Gaussian lacks support for intermediate granularity.}
\label{compare_main}
\end{figure*}

In Figure~\ref{compare_main}, we qualitatively evaluate the segmentation performance of different models on novel views.
Our model achieves high-quality segmentation, particularly under fine-grained settings---for example, it produces more precise contours for the egg yolk. For semi-transparent objects such as the glass cup in the first row and the plastic box in the bottom three rows, baseline models exhibit foreground-background confusion. In contrast, our model benefits from opacity fine-tuning and delivers superior segmentation quality in these challenging cases.
Moreover, our model achieves more stable control over segmentation granularity.
In the panoptic segmentation results, baseline models show poor sensitivity to granularity and even exhibit inconsistent or confused segmentations. By comparison, our model accurately controls segmentation granularity and exhibits smooth transitions across different levels. 
Finally, as shown in the finest-grained segmentation results in the last row, only our model successfully segments subtle object parts, such as the beak of bird, the duck eye, bunny ears, and the recessed area on the barrel---highlighting the significant advantage of Binary-Gaussian in fine-grained segmentation and validating the effectiveness of our progressive contrastive learning strategy.
More visualization examples are provided in Appendix B.

\begin{figure}[!b]
\centering
\includegraphics[width=\linewidth]{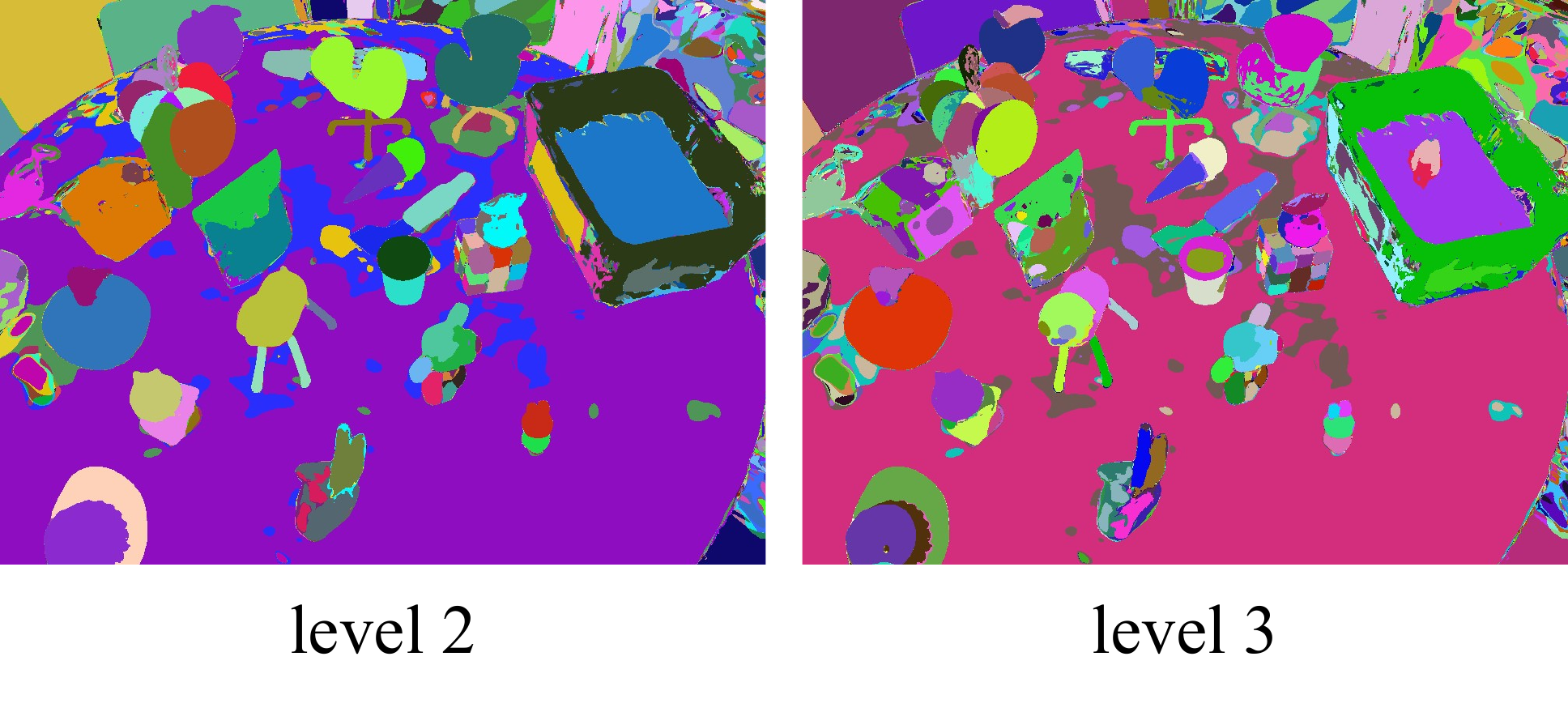} 
\caption{Segmentation results without the virtual negative across different granularities.}
\label{virtual_negative}
\end{figure}

\subsection{Ablation Study}
The ablation results for the virtual negative, opacity fine-tuning, and mask-balanced sampling are presented in Table~\ref{tab:ablation}.
\begin{table}[t]
\centering
\small
\begin{tabular}{l|cc}
\toprule
& \multicolumn{2}{c}{\textbf{mIoU (\%)}} \\
& Coarse & Fine \\
\midrule
Full      & 87.1 & 86.2 \\
\midrule
w/o Virtual Negative       & 85.3 & 77.9 \\
w/o Opacity Fine-tuning       & 86.0 & 84.4 \\
w/o MB Sampling      & 84.5 & 76.4 \\
\bottomrule
\end{tabular}
\caption{Ablation results for multi-granularity segmentation}
\label{tab:ablation}
\end{table}

\paragraph{Virtual Negative.}
Virtual negative is introduced to enhance consistency optimization for fine-grained segmentation, and removing this strategy leads to a significant drop in fine-grained segmentation performance. Figure~\ref{virtual_negative} presents a qualitative comparison, where we observe that for masks with limited granularity (e.g., table), the absence of virtual negatives causes the features to fail to form consistent representations at fine-grained levels.

\begin{figure}[!ht]
\centering
\includegraphics[width=\linewidth]{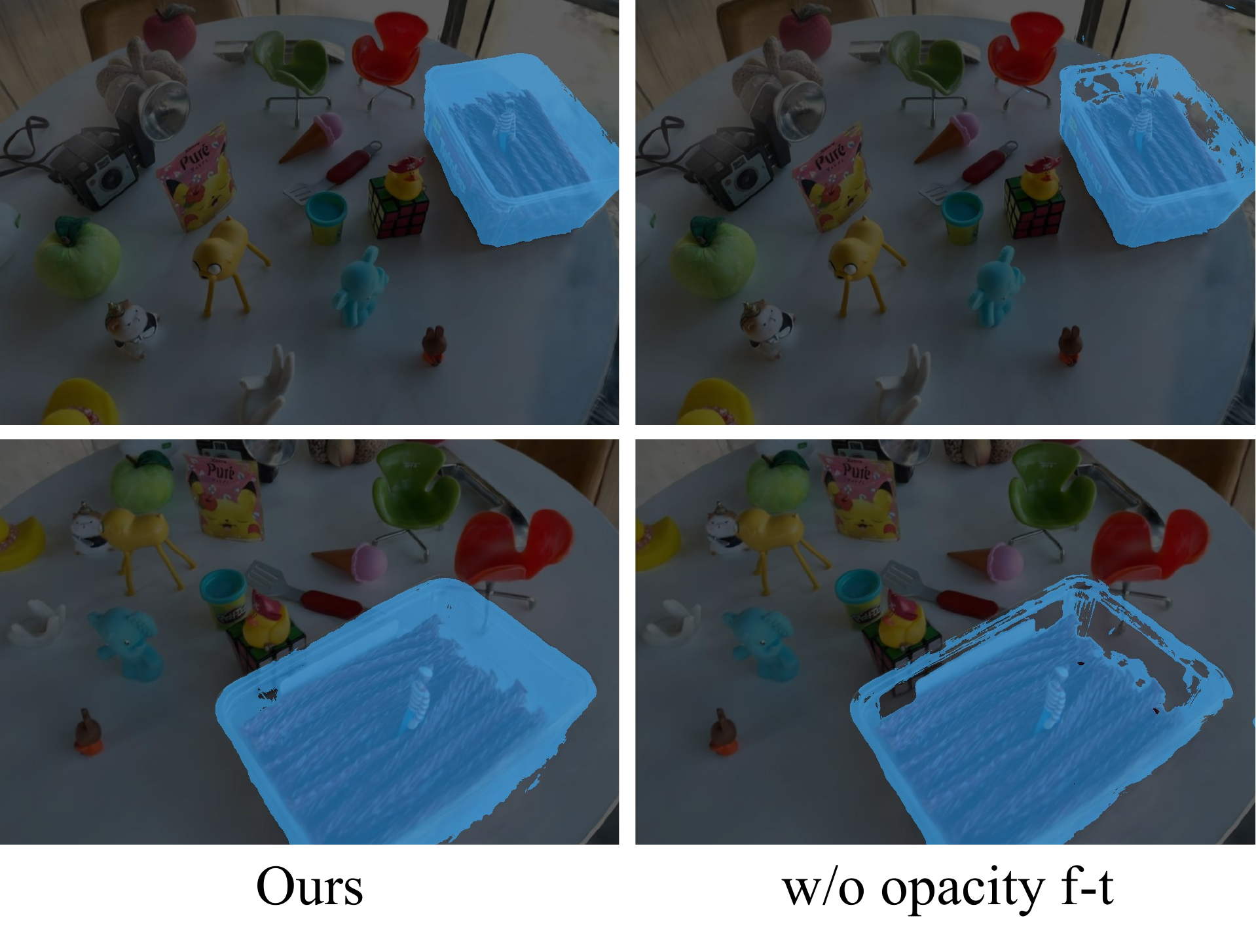} 
\caption{Ablation study on opacity fine-tuning for semi-transparent object segmentation.}
\label{opacity_finetuning}
\end{figure}
\paragraph{Opacity Fine-tuning.}
Since widely used benchmark datasets rarely contain objects with high transparency, opacity fine-tuning does not lead to significant improvements in quantitative metrics. To better demonstrate its effect, we provide a visual comparison in Figure~\ref{opacity_finetuning}, where opacity fine-tuning noticeably improves the segmentation quality of semi-transparent objects.

\paragraph{Mask-Balanced Sampling.}
We adopt Mask-Balanced Sampling to address the under-sampling issue of small-sized masks. As shown in Table~\ref{tab:ablation}, this strategy brings limited improvements to coarse segmentation but leads to notable gains in fine-grained segmentation, where the masks are typically smaller in size.

\section{Conclusion}
We propose a coarse-to-fine binary feature encoding scheme for 3D-GS-based segmentation, along with a progressive contrastive learning strategy tailored to this task. To address various challenges in this domain, we further introduce three training techniques: virtual negative, opacity fine-tuning, and mask-balanced sampling. Finally, with minimal parameter overhead and high inference speed, our method achieves excellent segmentation performance, especially at fine-grained levels.

\section*{Acknowledgements}
This work was supported by the National Natural Science Foundation of China under Grant No. U25A20409.

\bibliography{aaai2026}

@Article{kerbl3Dgaussians,
      author       = {Kerbl, Bernhard and Kopanas, Georgios and Leimk{\"u}hler, Thomas and Drettakis, George},
      title        = {3D Gaussian Splatting for Real-Time Radiance Field Rendering},
      journal      = {ACM Transactions on Graphics},
      number       = {4},
      volume       = {42},
      month        = {July},
      year         = {2023},
      url          = {https://repo-sam.inria.fr/fungraph/3d-gaussian-splatting/}
}

@article{fan2024lightgaussian,
  title={Lightgaussian: Unbounded 3d gaussian compression with 15x reduction and 200+ fps},
  author={Fan, Zhiwen and Wang, Kevin and Wen, Kairun and Zhu, Zehao and Xu, Dejia and Wang, Zhangyang and others},
  journal={Advances in neural information processing systems},
  volume={37},
  pages={140138--140158},
  year={2024}
}

@inproceedings{lee2024compact,
  title={Compact 3d gaussian representation for radiance field},
  author={Lee, Joo Chan and Rho, Daniel and Sun, Xiangyu and Ko, Jong Hwan and Park, Eunbyung},
  booktitle={Proceedings of the IEEE/CVF Conference on Computer Vision and Pattern Recognition},
  pages={21719--21728},
  year={2024}
}

@inproceedings{niedermayr2024compressed,
  title={Compressed 3d gaussian splatting for accelerated novel view synthesis},
  author={Niedermayr, Simon and Stumpfegger, Josef and Westermann, R{\"u}diger},
  booktitle={Proceedings of the IEEE/CVF Conference on Computer Vision and Pattern Recognition},
  pages={10349--10358},
  year={2024}
}

@article{navaneet2023compact3d,
  title={CompGS: Smaller and Faster Gaussian Splatting with Vector Quantization},
  author={Navaneet, KL and Meibodi, Kossar Pourahmadi and Koohpayegani, Soroush Abbasi and Pirsiavash, Hamed},
  journal={ECCV},
  year={2024}
}

@inproceedings{chen2024hac,
  title={Hac: Hash-grid assisted context for 3d gaussian splatting compression},
  author={Chen, Yihang and Wu, Qianyi and Lin, Weiyao and Harandi, Mehrtash and Cai, Jianfei},
  booktitle={European Conference on Computer Vision},
  pages={422--438},
  year={2024},
  organization={Springer}
}

@article{papantonakis2024reducing,
  title={Reducing the memory footprint of 3d gaussian splatting},
  author={Papantonakis, Panagiotis and Kopanas, Georgios and Kerbl, Bernhard and Lanvin, Alexandre and Drettakis, George},
  journal={Proceedings of the ACM on Computer Graphics and Interactive Techniques},
  volume={7},
  number={1},
  pages={1--17},
  year={2024},
  publisher={ACM New York, NY, USA}
}

@article{kobayashi2022decomposing,
  title={Decomposing nerf for editing via feature field distillation},
  author={Kobayashi, Sosuke and Matsumoto, Eiichi and Sitzmann, Vincent},
  journal={Advances in neural information processing systems},
  volume={35},
  pages={23311--23330},
  year={2022}
}

@inproceedings{tschernezki2022neural,
  title={Neural feature fusion fields: 3d distillation of self-supervised 2d image representations},
  author={Tschernezki, Vadim and Laina, Iro and Larlus, Diane and Vedaldi, Andrea},
  booktitle={2022 International Conference on 3D Vision (3DV)},
  pages={443--453},
  year={2022},
  organization={IEEE}
}

@inproceedings{goel2023interactive,
  title={Interactive segmentation of radiance fields},
  author={Goel, Rahul and Sirikonda, Dhawal and Saini, Saurabh and Narayanan, PJ},
  booktitle={Proceedings of the IEEE/CVF Conference on Computer Vision and Pattern Recognition},
  pages={4201--4211},
  year={2023}
}

@inproceedings{caron2021emerging,
  title={Emerging properties in self-supervised vision transformers},
  author={Caron, Mathilde and Touvron, Hugo and Misra, Ishan and J{\'e}gou, Herv{\'e} and Mairal, Julien and Bojanowski, Piotr and Joulin, Armand},
  booktitle={Proceedings of the IEEE/CVF international conference on computer vision},
  pages={9650--9660},
  year={2021}
}

@inproceedings{kerr2023lerf,
  title={Lerf: Language embedded radiance fields},
  author={Kerr, Justin and Kim, Chung Min and Goldberg, Ken and Kanazawa, Angjoo and Tancik, Matthew},
  booktitle={Proceedings of the IEEE/CVF International Conference on Computer Vision},
  pages={19729--19739},
  year={2023}
}

@inproceedings{kirillov2023segment,
  title={Segment anything},
  author={Kirillov, Alexander and Mintun, Eric and Ravi, Nikhila and Mao, Hanzi and Rolland, Chloe and Gustafson, Laura and Xiao, Tete and Whitehead, Spencer and Berg, Alexander C and Lo, Wan-Yen and others},
  booktitle={Proceedings of the IEEE/CVF international conference on computer vision},
  pages={4015--4026},
  year={2023}
}

@inproceedings{zhou2024feature,
  title={Feature 3dgs: Supercharging 3d gaussian splatting to enable distilled feature fields},
  author={Zhou, Shijie and Chang, Haoran and Jiang, Sicheng and Fan, Zhiwen and Zhu, Zehao and Xu, Dejia and Chari, Pradyumna and You, Suya and Wang, Zhangyang and Kadambi, Achuta},
  booktitle={Proceedings of the IEEE/CVF Conference on Computer Vision and Pattern Recognition},
  pages={21676--21685},
  year={2024}
}

@article{chen2023interactive,
  title={Interactive segment anything nerf with feature imitation},
  author={Chen, Xiaokang and Tang, Jiaxiang and Wan, Diwen and Wang, Jingbo and Zeng, Gang},
  journal={arXiv preprint arXiv:2305.16233},
  year={2023}
}

@article{bhalgat2023contrastive,
  title={Contrastive lift: 3d object instance segmentation by slow-fast contrastive fusion},
  author={Bhalgat, Yash and Laina, Iro and Henriques, Joao F and Zisserman, Andrew and Vedaldi, Andrea},
  journal={arXiv preprint arXiv:2306.04633},
  year={2023}
}

@article{fan2022nerf,
  title={Nerf-sos: Any-view self-supervised object segmentation on complex scenes},
  author={Fan, Zhiwen and Wang, Peihao and Jiang, Yifan and Gong, Xinyu and Xu, Dejia and Wang, Zhangyang},
  journal={arXiv preprint arXiv:2209.08776},
  year={2022}
}

@article{cen2023segment,
  title={Segment anything in 3d with nerfs},
  author={Cen, Jiazhong and Zhou, Zanwei and Fang, Jiemin and Shen, Wei and Xie, Lingxi and Jiang, Dongsheng and Zhang, Xiaopeng and Tian, Qi and others},
  journal={Advances in Neural Information Processing Systems},
  volume={36},
  pages={25971--25990},
  year={2023}
}

@inproceedings{ren2022neural,
  title={Neural volumetric object selection},
  author={Ren, Zhongzheng and Agarwala, Aseem and Russell, Bryan and Schwing, Alexander G and Wang, Oliver},
  booktitle={Proceedings of the IEEE/CVF Conference on Computer Vision and Pattern Recognition},
  pages={6133--6142},
  year={2022}
}

@inproceedings{gaussian_grouping,
    title={Gaussian Grouping: Segment and Edit Anything in 3D Scenes},
    author={Ye, Mingqiao and Danelljan, Martin and Yu, Fisher and Ke, Lei},
    booktitle={ECCV},
    year={2024}
}

@article{ying2023omniseg3d,
  title={OmniSeg3D: Omniversal 3D Segmentation via Hierarchical Contrastive Learning},
  author={Ying, Haiyang and Yin, Yixuan and Zhang, Jinzhi and Wang, Fan and Yu, Tao and Huang, Ruqi and Fang, Lu},
  journal={arXiv preprint arXiv:2311.11666},
  year={2023}
}

@inproceedings{kim2024garfield,
  title={Garfield: Group anything with radiance fields},
  author={Kim, Chung Min and Wu, Mingxuan and Kerr, Justin and Goldberg, Ken and Tancik, Matthew and Kanazawa, Angjoo},
  booktitle={Proceedings of the IEEE/CVF Conference on Computer Vision and Pattern Recognition},
  pages={21530--21539},
  year={2024}
}

@inproceedings{choi2024click,
  title={Click-gaussian: Interactive segmentation to any 3d gaussians},
  author={Choi, Seokhun and Song, Hyeonseop and Kim, Jaechul and Kim, Taehyeong and Do, Hoseok},
  booktitle={European Conference on Computer Vision},
  pages={289--305},
  year={2024},
  organization={Springer}
}

@inproceedings{cen2025segment,
  title={Segment any 3d gaussians},
  author={Cen, Jiazhong and Fang, Jiemin and Yang, Chen and Xie, Lingxi and Zhang, Xiaopeng and Shen, Wei and Tian, Qi},
  booktitle={Proceedings of the AAAI Conference on Artificial Intelligence},
  volume={39},
  number={2},
  pages={1971--1979},
  year={2025}
}

@inproceedings{shen2024flashsplat,
  title={Flashsplat: 2d to 3d gaussian splatting segmentation solved optimally},
  author={Shen, Qiuhong and Yang, Xingyi and Wang, Xinchao},
  booktitle={European Conference on Computer Vision},
  pages={456--472},
  year={2024},
  organization={Springer}
}

@inproceedings{jiang2024vr,
  title={Vr-gs: A physical dynamics-aware interactive gaussian splatting system in virtual reality},
  author={Jiang, Ying and Yu, Chang and Xie, Tianyi and Li, Xuan and Feng, Yutao and Wang, Huamin and Li, Minchen and Lau, Henry and Gao, Feng and Yang, Yin and others},
  booktitle={ACM SIGGRAPH 2024 Conference Papers},
  pages={1--1},
  year={2024}
}

@article{kerbl20233d,
  title={3d gaussian splatting for real-time radiance field rendering.},
  author={Kerbl, Bernhard and Kopanas, Georgios and Leimk{\"u}hler, Thomas and Drettakis, George},
  journal={ACM Trans. Graph.},
  volume={42},
  number={4},
  pages={139--1},
  year={2023}
}

@inproceedings{zhao2025isegman,
  title={iSegMan: Interactive Segment-and-Manipulate 3D Gaussians},
  author={Zhao, Yian and Xu, Wanshi and Zheng, Ruochong and Qiao, Pengchong and Liu, Chang and Chen, Jie},
  booktitle={Proceedings of the Computer Vision and Pattern Recognition Conference},
  pages={661--670},
  year={2025}
}

@article{yan20243dsceneeditor,
  title={3dsceneeditor: Controllable 3d scene editing with gaussian splatting},
  author={Yan, Ziyang and Li, Lei and Shao, Yihua and Chen, Siyu and Wu, Zongkai and Hwang, Jenq-Neng and Zhao, Hao and Remondino, Fabio},
  journal={arXiv preprint arXiv:2412.01583},
  year={2024}
}

@article{qu2025drag,
  title={Drag Your Gaussian: Effective Drag-Based Editing with Score Distillation for 3D Gaussian Splatting},
  author={Qu, Yansong and Chen, Dian and Li, Xinyang and Li, Xiaofan and Zhang, Shengchuan and Cao, Liujuan and Ji, Rongrong},
  journal={arXiv preprint arXiv:2501.18672},
  year={2025}
}

@inproceedings{mirzaei2023spin,
  title={Spin-nerf: Multiview segmentation and perceptual inpainting with neural radiance fields},
  author={Mirzaei, Ashkan and Aumentado-Armstrong, Tristan and Derpanis, Konstantinos G and Kelly, Jonathan and Brubaker, Marcus A and Gilitschenski, Igor and Levinshtein, Alex},
  booktitle={Proceedings of the IEEE/CVF Conference on Computer Vision and Pattern Recognition},
  pages={20669--20679},
  year={2023}
}

@inproceedings{fridovich2022plenoxels,
  title={Plenoxels: Radiance fields without neural networks},
  author={Fridovich-Keil, Sara and Yu, Alex and Tancik, Matthew and Chen, Qinhong and Recht, Benjamin and Kanazawa, Angjoo},
  booktitle={Proceedings of the IEEE/CVF conference on computer vision and pattern recognition},
  pages={5501--5510},
  year={2022}
}

@article{knapitsch2017tanks,
  title={Tanks and temples: Benchmarking large-scale scene reconstruction},
  author={Knapitsch, Arno and Park, Jaesik and Zhou, Qian-Yi and Koltun, Vladlen},
  journal={ACM Transactions on Graphics (ToG)},
  volume={36},
  number={4},
  pages={1--13},
  year={2017},
  publisher={ACM New York, NY, USA}
}

@article{mildenhall2019local,
  title={Local light field fusion: Practical view synthesis with prescriptive sampling guidelines},
  author={Mildenhall, Ben and Srinivasan, Pratul P and Ortiz-Cayon, Rodrigo and Kalantari, Nima Khademi and Ramamoorthi, Ravi and Ng, Ren and Kar, Abhishek},
  journal={ACM Transactions on Graphics (ToG)},
  volume={38},
  number={4},
  pages={1--14},
  year={2019},
  publisher={ACM New York, NY, USA}
}

@article{mildenhall2021nerf,
  title={Nerf: Representing scenes as neural radiance fields for view synthesis},
  author={Mildenhall, Ben and Srinivasan, Pratul P and Tancik, Matthew and Barron, Jonathan T and Ramamoorthi, Ravi and Ng, Ren},
  journal={Communications of the ACM},
  volume={65},
  number={1},
  pages={99--106},
  year={2021},
  publisher={ACM New York, NY, USA}
}

@inproceedings{yen2022nerf,
  title={Nerf-supervision: Learning dense object descriptors from neural radiance fields},
  author={Yen-Chen, Lin and Florence, Pete and Barron, Jonathan T and Lin, Tsung-Yi and Rodriguez, Alberto and Isola, Phillip},
  booktitle={2022 international conference on robotics and automation (ICRA)},
  pages={6496--6503},
  year={2022},
  organization={IEEE}
}

@article{grill2020bootstrap,
  title={Bootstrap your own latent-a new approach to self-supervised learning},
  author={Grill, Jean-Bastien and Strub, Florian and Altch{\'e}, Florent and Tallec, Corentin and Richemond, Pierre and Buchatskaya, Elena and Doersch, Carl and Avila Pires, Bernardo and Guo, Zhaohan and Gheshlaghi Azar, Mohammad and others},
  journal={Advances in neural information processing systems},
  volume={33},
  pages={21271--21284},
  year={2020}
}

@inproceedings{chen2021exploring,
  title={Exploring simple siamese representation learning},
  author={Chen, Xinlei and He, Kaiming},
  booktitle={Proceedings of the IEEE/CVF conference on computer vision and pattern recognition},
  pages={15750--15758},
  year={2021}
}

@article{cha2023regularizing,
  title={Regularizing with pseudo-negatives for continual self-supervised learning},
  author={Cha, Sungmin and Cho, Kyunghyun and Moon, Taesup},
  journal={arXiv preprint arXiv:2306.05101},
  year={2023}
}

\end{document}